\documentclass[10pt, journal]{IEEEtran}

\usepackage{amsmath,amssymb,amsfonts}
\usepackage{graphicx}
\usepackage{pifont}

\usepackage{amssymb}
\usepackage{amsthm}
\usepackage{xcolor}
\usepackage[figuresright]{rotating}

\usepackage{graphicx}
\graphicspath{{/}{fig/}}

\usepackage{comment}
\usepackage{array}
\usepackage{textcomp}
\usepackage{gensymb}
\usepackage{xcolor}
\usepackage{multirow}
\usepackage{booktabs}
\usepackage{enumitem}

\usepackage{mathtools}
\usepackage{breqn}
\usepackage{float}
\usepackage[noend]{algpseudocode}
\usepackage[vlined, ruled, shortend]{algorithm2e}

\usepackage{caption}
\usepackage{subcaption}

\usepackage{mathtools}
\usepackage{float}
\usepackage{hyperref}

\usepackage{pgfplots}
\pgfplotsset{compat=1.7}
\usepgfplotslibrary{groupplots}

\usepackage{multirow,tabularx}
\usepackage[ancient]{flushend}

\usepackage{xcolor}
\colorlet{orange_maxim}{green!10!orange!90!}

\usepackage[capitalise]{cleveref}

\newcommand{\rtoev}{\mathbb{R}^{\lvert \mathcal{E} \lvert \lvert \mathcal{V} \lvert}}

\DeclareMathOperator*{\argmin}{argmin}

\newlength\figureheight
\newlength\figurewidth
\usetikzlibrary{shapes.misc}

\tikzset{cross/.style={cross out, draw=black, minimum size=2*(#1-\pgflinewidth), inner sep=0pt, outer sep=0pt},
cross/.default={1pt}}

\renewcommand{\baselinestretch}{0.985}

\IEEEaftertitletext{\vspace{-1.5\baselineskip}}

\title{
    \huge 
    VIO-UWB-Based Collaborative Localization and Dense Scene Reconstruction within Heterogeneous Multi-Robot Systems
}

\author{
    Jorge Peña Queralta, Li Qingqing, Fabrizio Schiano,
    Tomi Westerlund
    \thanks{Jorge Peña Queralta, Li Qingqing and Tomi Westerlund are with the \href{https://tiers.utu.fi}{Turku Intelligent Embedded and Robotic Systems (TIERS) Lab, University of Turku, Finland}, e-mails: \{jopequ, qingqli, tovewe\}@utu.fi. Fabrizio Schiano is with the \href{https://www.epfl.ch/labs/lis/}{Laboratory of Intelligent Systems (LIS)}, École Polytechnique Fédérale de Lausanne (EPFL), Switzerland. Email: \{fabrizio.schiano\}@epfl.ch}
}

\begin{document}

\maketitle
\thispagestyle{empty}
\pagestyle{empty}


\begin{abstract}

    
    Effective collaboration in multi-robot systems requires accurate and robust estimation of relative localization: from cooperative manipulation to collaborative sensing, and including cooperative exploration or cooperative transportation. This paper introduces a novel approach to collaborative localization for dense scene reconstruction in heterogeneous multi-robot systems comprising ground robots and micro-aerial vehicles (MAVs). We solve the problem of full relative pose estimation without sliding time windows by relying on UWB-based ranging and Visual Inertial Odometry (VIO)-based egomotion estimation for localization, while exploiting lidars onboard the ground robots for full relative pose estimation in a single reference frame. During operation, the rigidity eigenvalue provides feedback to the system. To tackle the challenge of path planning and obstacle avoidance of MAVs in GNSS-denied environments, we maintain line-of-sight between ground robots and MAVs. Because lidars capable of dense reconstruction have limited FoV, this introduces new constraints to the system. Therefore, we propose a novel formulation with a variant of the Dubins multiple traveling salesman problem with neighborhoods (DMTSPN) where we include constraints related to the limited FoV of the ground robots. Our approach is validated with simulations and experiments with real robots for the different parts of the system.
    
    

\end{abstract}



\IEEEpeerreviewmaketitle





\section{Introduction}\label{sec:introduction}

Research in multi-robot systems and swarm robotics has seen increasing attention from the research community in recent years~\cite{hayat2016survey}. Swarms of robots and the robots' ability to navigate and explore unknown and extreme environments have been identified as two of the grand challenges in robotics for the next decade~\cite{yang2018grand}. Heterogeneous multi-robot systems and algorithms for collaborative autonomy have also gained increasing research interest owing to the potential for multi-modal and multi-source sensor fusion, and the deployment flexibility and robustness in complex scenarios~\cite{rizk2019cooperative, queralta2020collaborative}.

\begin{figure}
    \centering
    \begin{subfigure}[t]{0.48\textwidth}
        \includegraphics[width=\textwidth]{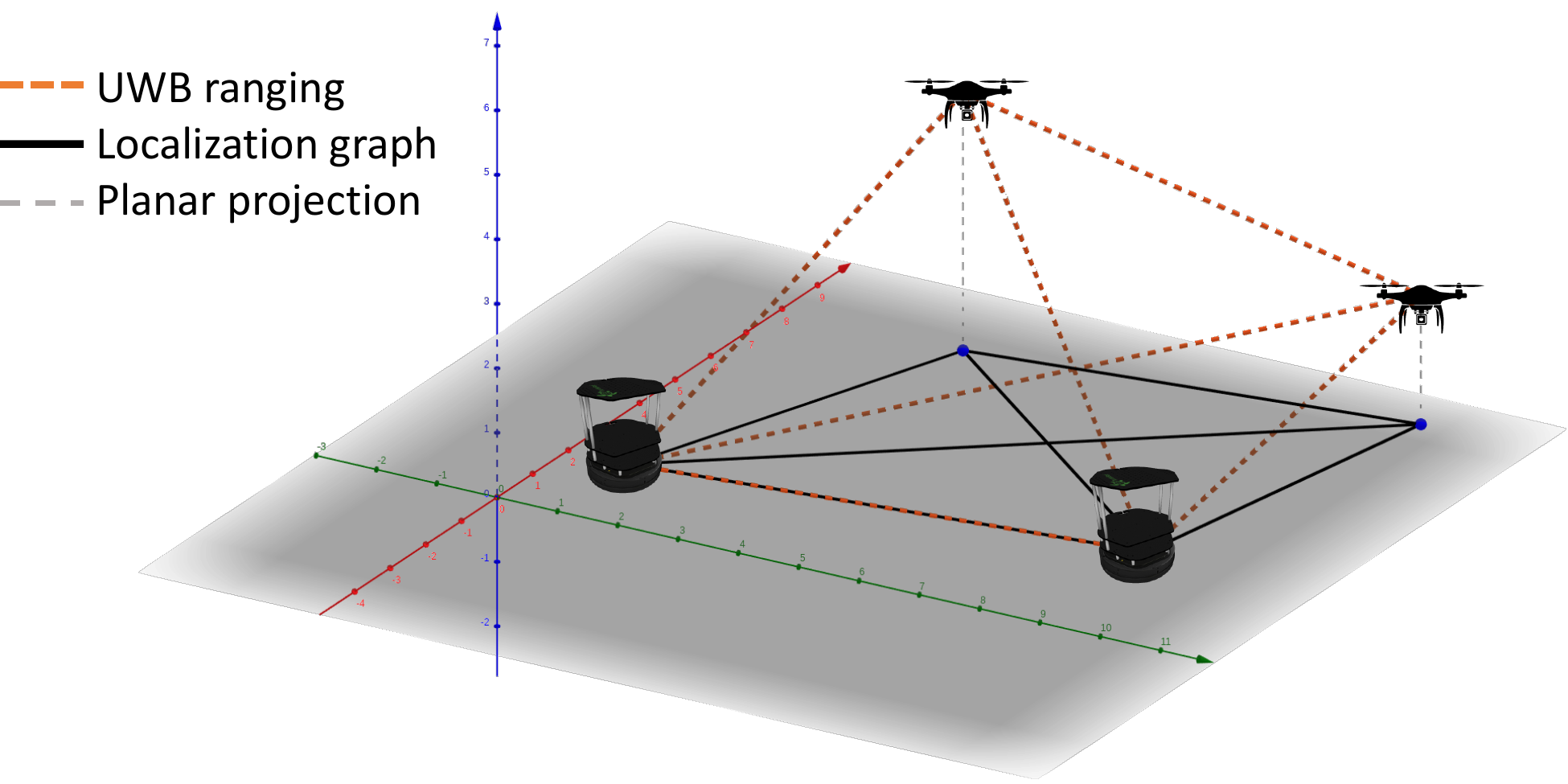}
        \caption{Conceptual illustration of collaborative localization in a heterogeneous multi-robot system where the localization graph is globally rigid.}
        \label{subfig:projection}
    \end{subfigure}
    \begin{subfigure}[t]{0.48\textwidth}
        \includegraphics[width=\textwidth]{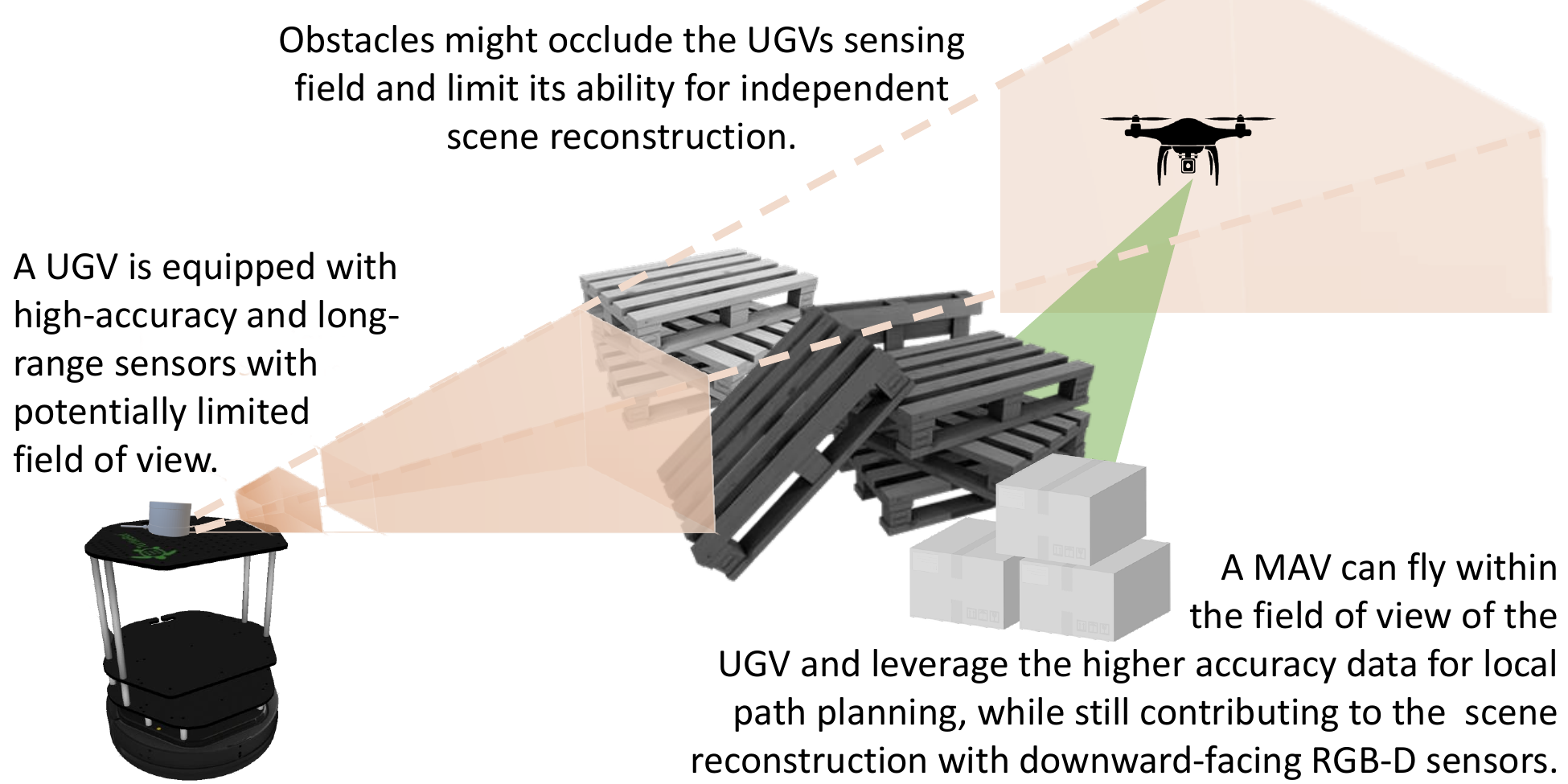}
        \caption{Conceptual illustration of collaborative scene reconstruction with a heterogeneous UGV+MAV multi-robot system.}
        \label{subfig:sensing}
    \end{subfigure}
    \caption{In this paper, we explore the problem of UWB-based collaborative localization based on graph rigidity (a) for collaborative sensing (b) and dense scene reconstruction.}
    \label{fig:concept}
    \vspace{0.23em}
\end{figure}

In GNSS-denied environments, different approaches to multi-robot cooperative exploration have been showcased during the DARPA Subterranean challenge~\cite{rouvcek2019darpa, petrlik2020robust}. The participants of the challenge deployed both unmanned ground vehicles (UGVs) and micro aerial vehicles (MAVs). Localization and collaborative sensing represented two of the main challenges. Indeed, solving these problems would allow robotic systems to go out of laboratory settings. In this paper, we address the problem of collaborative multi-robot 3D dense scene reconstruction involving UGVs and MAVs.

Solid-state lidars represent the state-of-the-art regarding sensors for high-accuracy and long-range dense point cloud scanners, often with limited Field of View (FoV) owing to the lack of rotating parts~\cite{li2020towards, lin2020loam}. 
 However, the stringent payload constraints of MAVs 
make RGB-D cameras a more viable solution for dense scene reconstruction~\cite{huang2017visual}. Therefore, to achieve collaborative sensing, we chose to embed 3D lidars only on UGVs with MAVs equipped with depth cameras.

From the point of view of MAVs, advances in both monocular and stereo dense reconstruction and navigation have arrived to a point where commercial solutions such as Skydio II are able of high degrees of autonomy and situational awareness~\cite{ackerman2019skydio}. Nonetheless, deployment in GNSS-denied environments with limited visibility and potentially dynamic environments is still challenging~\cite{shakhatreh2019unmanned}. 

Regarding localization approaches for multi-robot systems in GNSS-denied environments, ultra-wideband (UWB) wireless ranging transceivers have recently emerged as an inexpensive and relatively accurate method for point-to-point ranging~\cite{queralta2020uwbbased, shule2020uwb}. Full pose estimation can be achieved by fusing UWB with visual-inertial odometry (VIO)-based egomotion estimation. However, previous works are not able to provide full position estimation within a single common reference unless robots move~\cite{xu2020decentralized}.

At the same time, over the past two decades research in both distance-based~\cite{zelazo2012rigidity} and bearing-based~\cite{schiano2016rigidity, schiano2017bearing,schiano2018dynamic} rigidity maintenance control have been shown to be robust and efficient methods for distributed collaborative localization in multi-robot systems. Different rigidity maintenance approaches have been proposed to ensure that the collaborative localization problem can be solved. However, up to our knowledge, none of the works of the literature integrate rigidity theory as a feedback to a localization system while, at the same time, achieving a navigation goal given by a high-level planner.

Taking the above considerations into account, we present a novel approach to collaborative dense scene reconstruction within heterogeneous multi-robot systems that address several of the aforementioned challenges. First, we propose a UWB-VIO-based collaborative localization framework that exploits sensors onboard UGVs to detect the position of MAVs at startup for a unique localization graph realization. Second, we leverage the UGVs' sensors high accuracy and range for real-time path planning and obstacle avoidance of MAVs. At the same time, we use the collaborative localization framework to extract data from the UGVs' point clouds relative to the MAVs' positions. Third, we propose a different formulation to the collaborative scene reconstruction problem that solves a variant of the Dubins Multiple Traveling Salesman Problem with Neighborhoods (DMTSPN). In this problem the neighborhoods are defined as the parts of the environment which are occluded, due to obstacles, for the UGVs' sensors. Finally, to account for the limited FoV of lidars, we solve the DMTSPN by adding UGV-to-MAV line-of-sight (LoS) constraints.

Compared to previous approaches in rigidity maintenance for collaborative localization, we focus on path planning while ensuring LoS, which, in turn, ensures that the graph is rigid within the UWB range. Compared to previous works on dense scene reconstruction with map merging or point cloud alignment, we focus on teaming the different robots and the distribution of locations to be surveyed. Our results show good alignment of point clouds without further optimization, which can then be improved based on existing algorithms.

\section{Related Works}\label{sec:related_works}


\subsection{UWB-based collaborative localization}

Ultra-wideband localization systems often rely on time-of-flight measurements between moving nodes, or tags, and a set of fixed nodes, or anchors, in known positions~\cite{nguyen2018integrated, nguyen2019integrated, shule2020uwb}. These systems are more portable and inexpensive than motion capture (MOCAP) systems but trade-off accuracy, both lacking deployment flexibility. Recent approaches to mobile localization systems include continuous recalibration of anchor positions~\cite{almansa2020autocalibration}, or collaborative localization approaches. Xu et al. have presented a robust multi-modal sensor fusion algorithm exploiting UWB ranging and VIO that provides a decentralized and collaborative localization framework for multi-UAV systems~\cite{xu2020decentralized}. In other works, UWB ranging has helped overcome the limitations of lidar-based SLAM in environments lacking enough features (e.g., long corridors)~\cite{song2019uwb}.

Similar work in collaborative localization is being carried out by Walter et al. with UVDAR~\cite{walter2019uvdar}, an ultra-violet detection and ranging sensor that uses active markers and cameras onboard UAVs. However, this approach requires not only LoS but also the markers to be within the field of view of the cameras. One benefit compared to UWB ranging is that UVDAR provides full relative position information~\cite{walter2018fast}. With multiple UWB transceivers on at least one of the robots, relative position and not only ranging is also possible~\cite{nguyen2018robust}.

\subsection{Collaborative scene reconstruction}

Over the last decade we witnessed an increasing adoption of inexpensive 2D lidars, 3D scanners, and depth cameras on mobile robots. Also, several advancements have been presented in monocular dense SLAM. These aspects pave the way for the development of multiple approaches for cooperative mapping and collaborative scene reconstruction. 
Collaborative SLAM and collaborative mapping in general have been widely studied problems. However, the focus is often on the map merging or area coverage distribution among the robots~\cite{schmuck2017multi, mahdoui2019communicating}. Relevant to this work is collaborative RGB-D reconstruction~\cite{zollhofer2014real}. 

We take a different approach by focusing on path planning with constraints to visit a series of interest regions. In this direction, a recent work by Dong et al. on collaborative dense scene reconstruction that takes an initial map and focuses on task allocation is similar to ours from the formulation point of view~\cite{dong2019multi}. Another related area is next best view (NBV) planning. In~\cite{sukkar2019multi}, Sukkar et al. present a multi-robot region-of-interest reconstruction with RGB-D cameras. Compared to these approaches, we focus on the integration of sensing constraints (e.g., limited FoV) in LoS path planning.

\section{Collaborative Localization}\label{sec:collab_localiz}

We use the following notation for the remaining of this paper. We consider a group of $N$ robots, or agents, with positions denoted by $\textbf{p}_i(t)\in\mathbb{R}^3$, with $i\in\{1,\dots,N\}$. Agents are able to measure their relative distance to a subset of the other agents in a bidirectional way, i.e., with both agents calculating a common ranging estimation simultaneously. 
The set of robots and estimated distances between them are modeled by a graph $\mathcal{G}=(\mathcal{V},\mathcal{E})$, where $\{1,\dots,N\}$ is a set of $N$ vertices and $\mathcal{E} \subset \mathcal{V}\times \mathcal{V}$ is a set of $M \leq N(N-1)/2$ edges. We consider undirected graphs, i.e., $(i,j) \in \mathcal{E} \Longleftrightarrow (j,i) \in \mathcal{E}$.

We consider robots in three-dimensional space. Nonetheless, for the purpose of collaborative localization, we assume that the position of each agent is given by $\textbf{p}_i\in\mathbb{R}^2$. The pair $(\mathcal{G},\textbf{p})$ with the position vector $\textbf{p} = [\textbf{p}_1^T,\dots,\textbf{p}_N^T]^T$ is a framework. We denote the incidence, degree and adjacency matrices by $E(\mathcal{G}) \in \rtoev$, $\Lambda(\mathcal{G})\in\rtoev$ and $A(\mathcal{G}) \in \rtoev$, respectively, and the graph laplacian by $\mathcal{L}(\mathcal(G)) = E(\mathcal{G})E(\mathcal{G})^T=\Lambda(\mathcal{G})-A(\mathcal{G})$. It is worth noting that an important result from algebraic graph theory states that $\mathcal{G}$ is connected if and only if the second smallest eigenvalue of $L(\mathcal{G})$ is positive~\cite{godsil2013algebraic}.

The solution of the collaborative localization problem is related to the uniqueness of graph realizations in space. There will be a unique graph realization (except for rototranslations in $\mathbb{R}^2$) if the framework $(\mathcal{G}, \textbf{p})$ is rigid. Intuitively, a rigid graph is a graph that cannot be deformed without breaking the constraints put over the edges. A sufficient and necessary algebraic condition similar to that of the graph connectivity can be given. We first define an edge constraint function $g_{\mathcal{G}} : \rtoev \mapsto \mathbb{R}^{\lvert \mathcal{E} \lvert}$ that through $g_{\mathcal{G}}(\textbf{p})$ defines a constraint over each of the edges $g_{ij}(\textbf{p}_i,\textbf{p}_j) \forall (i,j) \in \mathcal{E}$. We will use $g_{ij}(\textbf{p}_i,\textbf{p}_j)=\lVert\textbf{p}_i-\textbf{p}_j\lVert^2$ to use distances as constraints.

Two frameworks that represent realizations of the same graph, $(G,\textbf{p}_1)$ and $(G,\textbf{p}_2)$, are said to be equivalent if $g_{\mathcal{G}}(\textbf{p}_1) = g_{\mathcal{G}}(\textbf{p}_2)$, and congruent if $g_{\mathbb{K}}(\textbf{p}_1) = g_{\mathbb{K}}(\textbf{p}_2)$, where $\mathbb{K}$ is the complete graph with the same vertex set $\mathcal{V}$ as $\mathcal{G}$. A framework is \textit{locally rigid} if $\forall \textbf{p}\in\rtoev \:\exists\mathcal{P}\subset\rtoev,\:\textbf{p}\in\mathcal{P}$ such that
$g^{-1}_{\mathcal{G}}\left( g_{\mathcal{G}}(\textbf{p}) \right) \cap \mathcal{P} = g^{-1}_{\mathbb{\mathbb{K}}}\left( g_{\mathbb{K}}(\textbf{p}) \right) \cap \mathcal{P}$ and \textit{globally rigid} if $\forall \textbf{p}\in\rtoev$,
$    g^{-1}_{\mathcal{G}}\left( g_{\mathcal{G}}(\textbf{p}) \right) = g^{-1}_{\mathbb{K}}\left( g_{\mathbb{K}}(\textbf{p}) \right)%
$.

Even if a locally rigid graph is achieved, the null space of the transformations is defined by rototranslations of the graph realization in $\mathbb{R}^2$. In our experiments, we will consider the position of one of the agents as the origin of coordinates of the system, and utilize sensors onboard the UGVs to establish the orientation of the graph realization, after rigidity is ensured and MAVs take off. Therefore, all the measurements are relative to the agent chosen as the origin. 

We now consider infinitesimal rigid frameworks as those where constraints are met under infinitesimal perturbations $\delta\textbf{p}$. In order to maintain the constraints over edges, we can compute the Jacobian matrix

\begin{equation}
 \dot{g}_{\mathcal{G}}\left(\textbf{p}(t)\right) = \textbf{0} \Longrightarrow \dfrac{\delta g_{\mathcal{G}(p)}}{\delta(\textbf{p})}\dot{\textbf{p}} = R_{\mathcal{G}}(\textbf{p})\dot{\textbf{p}} = \textbf{0}
\end{equation}

where $R_{\mathcal{G}}(\textbf{p}) \in \mathbb{R}^{d \lvert\mathcal{E}\rvert \lvert\mathcal{V}\rvert}$ is the rigidity matrix~\cite{zelazo2012rigidity}, with $d=2$ because we are considering points in $\mathbb{R}^2$. Translations and rotations in the Cartesian space make up the non-trivial kernel of $R_{\mathcal{\mathbb{K}}}(\textbf{p})$, and therefore we can say that a framework $(\mathcal{G},\textbf{p})$ is infinitesimally rigid if the rank of $R_{\mathcal{G}}(\textbf{p})$ is the same as that of $R_{\mathcal{\mathbb{K}}}(\textbf{p})$: $2N-3$ in $\mathbb{R}^2$ (equivalently, $3N-6$ in $\mathbb{R}^3$).

In this paper, we are considering a heterogeneous multi-robot system comprising both UGVs and MAVs, and therefore their relative positions must be given in a three-dimensional space. While the above conditions for rigidity hold for three-dimensional graphs, the amount of information required is larger (more edges needed in $\mathcal{G}$). Since MAVs are already equipped with relative altitude sensors and capable of VIO estimations, we only consider graph rigidity in 2D and project the ranging information to the plane using data from other onboard sensors. The relative altitude is estimated primarily based on a downward facing single-beam lidar sensor. 
We model UWB measurements with Gaussian noise:
\begin{equation}
    \textbf{z}^{UWB}_{(i,j),\:(i,j)\in\mathcal{E}} = \lVert \textbf{p}_i(t)-\textbf{p}_j(t) \lVert + \mathcal{N}(0,\sigma_{UWB})
\end{equation}
where $\sigma_{UWB}$ is obtained experimentally from our previous work~\cite{almansa2020autocalibration}. VIO egomotion estimations are modeled with
\begin{dmath}
    \textbf{z}^{VIO}_{i,\:i\in\mathcal{V}} = \left[ \begin{array}{cc} \textbf{R}_i(t-\delta t)\hat{\textbf{R}}_i(t) & \lVert \textbf{p}_i(t)-\textbf{p}_i(t-\delta t) \lVert \\ 0 & 1 \end{array} \right] + \mathcal{N}(0,\sigma_{VIO})
\end{dmath}
where we utilize $\sigma_{VIO}<\sigma_{UWB}/10$, estimated experimentally, and $\delta t$ is the output frequency of the VIO algorithm, $\textbf{R}_i(t)$ is the orientation matrix for agent $i$ and $\hat{\textbf{R}}_i(t)$ the relative egomotion estimation in the interval $(t-\delta t, t]$.
The relative altitude is estimated based on the lidar 
\begin{equation}
    \textbf{z}^{H}_{i,\:i\in\mathcal{V}} = \left\{ \begin{array}{ll} h^{lidar}_i &\text{if}\:\: \Delta(h^{lidar}_i, h^{UWB}_i,h^{VIO}_i) < 1 \\[+2pt] h^{UWB+VIO}_i & \text{otherwise} \end{array} \right.
\end{equation}
where $\Delta(\cdot)$ estimates the mismatch between the different sensors. We use a filter to \textit{smooth} the UWB ranges using the VIO translational estimations at both agents.

UWB-based ranging information provides only the position of the full pose of the agents. Therefore, we rely on VIO-based orientation to estimate the orientation and achieve full pose estimation. We assume that all agents share a common reference. However, in order to match the graph realization with the agents' reference, we need to be able to measure the relative position, and not just distance, of at least one pair of agents. To do so, we assume that each MAV is within at least one UGV's field of view when the mission starts, and that the MAV can be detected from the UGV after taking off. 
Let $\mathcal{V}_{UGV}$ and $\mathcal{V}_{MAV}$ be the sets of UGVs and MAVs, respectively, with $N=N_{UGV}+N_{MAV}=\lvert\mathcal{V}_{UGV}\lvert+\lvert\mathcal{V}_{MAV}\lvert$. Then, let $\mathcal{G}_S=(\mathcal{V}_S,\mathcal{E}_S)$ be the UGV sensing graph with the same vertex set $\mathcal{V}_S=\mathcal{V}$ as $\mathcal{G}$ where $(i,j)\in\mathcal{E}_S \Leftrightarrow i\in\mathcal{V}_{UGV}, \:j\in\mathcal{V}_{MAV}$ and $j$ is \textit{visible} from sensors onboard $i$ (e.g. 3D lidar). We denote by $\mathcal{N}_i$ the set of neighbor nodes in $\mathcal{G}$ and $\mathcal{N}_{S_i}$ the neighbor set in $\mathcal{G}_S$. The actual localization is done by minimizing triangulation errors.

We now assume that the sensors on the UGVs produce accurate and dense point clouds but with limited FoV. Let $\mathcal{P}_i\subset\mathbb{R}^3$ be the point cloud generated at agent $i$. For a point $p\in\mathbb{R}^3$, we denote by $\mathcal{N}_p^R(\mathcal{P})=\{q\in\mathcal{P}\mid \lVert p-q \lVert \leq R\}$ the set of points in $\mathcal{P}$ within a distance $R$ of $p$. Finally, given a graph realization orientation $\theta$, we denote by $\hat{p}_i(\theta)$ the estimated position of agent $i$ in the global reference frame, where the position of agent $0$ is used as the origin of the frame. We then calculate the orientation $\hat{\theta}$ of the graph realization in the common reference frame by minimizing :%
\begin{equation}
    \hat{\theta} = \argmin_{\theta} \displaystyle\sum_{i\in\mathcal{V}_{UGV}} \displaystyle\sum_{j\in\mathcal{N}_{S_i}} 
    \dfrac{\lvert \mathcal{N}_{\hat{p}_i(\theta)}^{R_{MAV}}\left(\mathcal{P}\right)\lvert}
    {\lvert\mathcal{N}_{\hat{p}_i(\theta)}^{2R_{MAV}}\left(\mathcal{P}\right)\lvert+1}
\end{equation}%
where $R_{MAV}$ is the radius of the circumscribed sphere to a MAV point cloud, roughly half the width of a MAV (we consider homogeneous MAVs, otherwise the different sizes must be taken into account). The localization process is summarized in Algorithm~\ref{alg:localization_algotirhm}.

    \begin{algorithm}[t]
    \small
	\caption{Collaborative localization.}
	\label{alg:localization_algotirhm}
	\KwIn{\\
	    \hspace{1em}UWB Ranges : $\{\textbf{z}^{UWB}_{(i,j)}\}\in\mathbb{R}^{\lvert\mathcal{E}\lvert}$; \\
	    \hspace{1em}3D lidar point cloud $\{\mathcal{P}_i\}$; \\
	    \hspace{1em}Relative altitude of MAVs: \{$\textbf{z}^H_i\}$; \\
	    \hspace{1em}VIO odometry: $\{\textbf{z}^{VIO}_i\}$ \\
	}
	\KwOut{\\
	    \hspace{1em} Full robot poses $\hat{\textbf{p}}(\hat{\theta})\in\mathbb{R}^{6}$
	}  
	\BlankLine
    \While {!\textit{graph\_is\_connected}$\left(L(\mathcal{G})\right)$} {
        sleep();\\
    }
    \While {!\textit{graph\_is\_rigid}$\left(R_{\mathcal{G}}(\textbf{z}^{UWB})\right)$} {
        sleep();\\
    }
    $\hat{\textbf{p}}\leftarrow\:minimize\_triangulation\_error();$\\
    takeoff\_MAVs();\\
    Calculate graph orientation $\hat{\theta}$ as follows (init. $\theta=0$): \\
    \While{not exit\_condition$(\hat{\theta})$} {
        $error=0; \:\:\:theta += \Delta\theta$\;
        \ForEach{$i \in \mathcal{V}_{UGV}$} {
            Generate K-D Tree from point cloud: $kdtree_i\leftarrow\mathcal{P}_i$;\\
            \ForEach{$j\in\mathcal{N}_{S_i}$} {
                $error += size\left(kdtree_i.radiusSearch(\hat{\textbf{p}}_i,\:R_{MAV})\right)/$\\
                $size\left(kdtree_i.radiusSearch(\hat{\textbf{p}}_i,\:2R_{MAV})\right)$
            }
        }
        \If{$error < threshold$} {
            $\hat{\theta} = \theta;$\\
        }
    }
    Calculate full pose:
    \hspace{1em}$\hat{\textbf{p}}(\hat{\theta}) \:|\: \hat{\textbf{p}}_0 = (0,0)$,\:$\hat{\textbf{p}}_1 = \left(0,\textbf{z}_{(0,1)}^{UWB}\right)$;\\
\end{algorithm}

While the assumption of having MAVs within UGVs field of view could be directly leveraged towards measuring relative positions (together with a common orientation reference and VIO estimations at each robot), we still rely on UWB as the main source of localization when the mission starts and during the entire mission. The reason for doing so is twofold: first, UWB ranging is more accurate than estimating the position of an object extracted from a point cloud accounting for its size, shape and orientation; and second, we do not need to consider uncertainties in the point-cloud-based detector (whether it detects MAVs or other similarly sized objects).


\section{Collaborative Scene Reconstruction}\label{sec:collab_sensing}

The main objective of this paper is to provide methods for collaborative dense scene reconstruction. This serves simultaneously as a validation of the collaborative localization framework owing to the unavailability of high-accuracy tracking systems such as those utilized in the \textit{CoLo} evaluation~\cite{chen2019colo}.

In order to formulate the problem, we use the following notation. Let $\textbf{q}\in\mathbb{R}^{2M}, \:\textbf{q}=[\textbf{q}_1^T,\dots,\textbf{q}_N^T]^T$ be a stacked position vector for the $M$ locations to be visited. We make the following assumptions: (i) $M\geq N_{MAV}\geq N_{UGV}$; and (ii) if $H(\textbf{p}_{UGV})$ is the convex hull defined by the positions of the UGVs, then $\textbf{q}\cap H(\textbf{p}_{UGV})=\emptyset$, i.e., all locations to be surveyed by the MAVs lay \textit{beyond} the positions of the UGVs.

We consider the problem of utilizing multiple MAVs to obtain information about the areas of the scene that are occluded to the UGVs, with each MAV always staying within the FoV of one UGV (its \textit{tracker}). The collaborative scene reconstruction process then proceeds as follows (see Algorithm~\ref{alg:sensing_algorithm}). First, the ground robots scan the scene  and estimate blind spots based on their movement constraints due to uneven terrain or near obstacles. Second, each of the occluded regions behind obstacles blocking the FoV of UGVs is considered a neighborhood for the DMTSPN problem. We distribute the locations to be surveyed among the MAVs by solving DMTSPN problem where each location represents a neighborhood. 

Multiple works have been devoted to solving DMTSPN~\cite{isaacs2011algorithms, isaacs2013dubins}. However, to the best of our knowledge, the current literature does not consider scenarios where the paths have to meet LoS with limited FoV constraints with respect to a certain point in the environment. In particular, considering limited FoV is a topic largely unaddressed in the multi-robot systems literature~\cite{soria2019influence}.

To address this, we assign each MAV to one of the UGVs based on their initial positions, and proceed to assign the neighborhoods to MAVs based on their angular position with respect to their \textit{tracker} UGV when considering their position in polar coordinates from the UGV's local reference. Then, the MTSPN problem is solved based on the constraints defined above, with the neighborhood assignment to MAVs changing until an exit condition is met. Finally, the Dubins paths are generated smoothing turns over the neighborhoods, and the speed of the MAVs is adjusted to ensure LoS maintenance (the DMTSPN solution does not consider time). 


\begin{algorithm}[t]
    \small
	\caption{Collaborative scene reconstruction.}
	\label{alg:sensing_algorithm}
	\KwIn{\\
	    \hspace{1em}Dense UGV lidar point clouds: $\{\mathcal{P}^{L}_{i}\}_{i\in \mathcal{V}_{UGV}}$ \\
	    \hspace{1em}Dense MAV depth point clouds: $\{\mathcal{P}^{D}_{i}\}_{i\in \mathcal{V}_{MAV}}$ \\
	    \hspace{1em}UGV FoV: $\{(\Delta\theta_i^H, \Delta\theta_i^V)\}_{i\in \mathcal{V}_{UGV}}$\\
	}
	\KwOut{\\
	    \hspace{1em} UGV+MAV paths $\{\hat{\textbf{p}}_i(t)\}$
	}  
	\BlankLine
	$\mathcal{P} = merge\_point\_clouds\left(\{\mathcal{P}_i^{L}\}\right);$\\
	$\mathcal{NBH} = neighborhoods\_from\_blind\_spots(\mathcal{P})$;\\
	$\{\mathcal{NBH_i}\}_{i\in\mathcal{V}_{MAV}} = assign\_nbh(\mathcal{NBH}, \textbf{p}_i^{MAV})$;\\
	\BlankLine
	\ForEach{$i \in \mathcal{V}_{UGV}$} {
	    \While{!neighborhood\_exit\_condition()} {
	        \ForEach{$j \in \mathcal{N}_{S_i}$} {
    	        $run\_tspn\_solver(N_i, \textbf{p}_i, \Delta\theta_i^H, \Delta\theta_i^V)$;\\
    	        $calculate\_dubins\_path(i);$\\
    	    }
	    }
        $merge\_while\_navigating(\{\mathcal{P}^{L}\}, \mathcal{P}^{D}\});$
    }
	    
            
\end{algorithm}

Since the different parts of the problem are decoupled (diving neighborhoods among MAVs, solving the TSPN for each MAV, and \textit{smoothing} the paths with Dubns curves), we will obtain a sub-optimal solution. However, this already happens when introducing the LoS with limited FoV constraints, since in general optimal solutions to the MTSPN will not ensure that MAVs can stay within LoS of UGVs. Because our focus is on providing an initial approach that also combines the collaborative localization framework, we leave the optimization of the DMTSP solution to future works.

\section{Methodology}\label{sec:methodology}



\begin{figure}[t]
    \centering
    \includegraphics[width=0.45\textwidth]{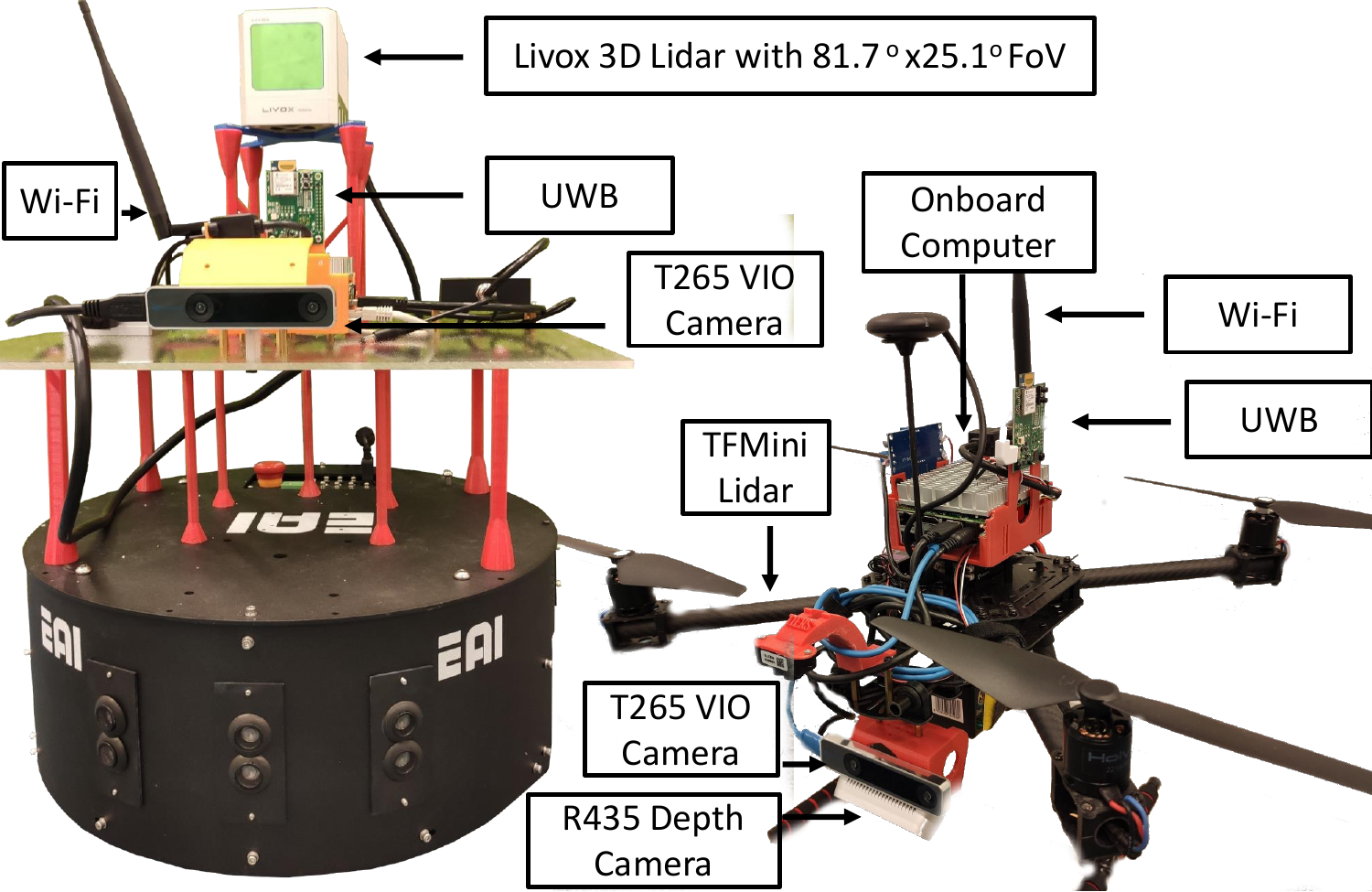}
    \caption{UGV and MAV utilized in the experiments.}
    \label{fig:robots}
\end{figure}

\textbf{Heterogeneous Multi-Robot System.} In our experiments, we utilize a single ground robot, one MAV, and a set of UWB transceivers (Fig.~\ref{fig:robots}). The ground robot is an EAI Dashgo platform equipped with a Livox Horizon lidar ($81.7\degree\times25.1\degree$ FoV). We also use one custom-built MAV based on the X500 quadrotor frame. The MAV is embedded with a Pixhawk flight controller running the PX4 firmware. A TF Mini Lidar is utilized for height estimation on the MAV, also equipped with Intel RealSense D435 depth camera for sensing and 3D reconstruction. 
An AAEON Up Square with dual-code Intel Celeron processor is used as a companion computer on both robots.
Both robots use RealSense T265 cameras for VIO-based egomotion estimation. 

\textbf{UWB Ranging and Position Estimation.} The distance between each pair of robots is estimated using Decawave DWM1001 UWB transceivers. In order to ensure a safe fallback for the MAV autonomous flight, we utilize both an anchor-based localization system as well as ranging between robots for collaborative localization. We also compare the localization from both methods.



\textbf{Software.} The system has been implemented using ROS Melodic, with all robots running the same version under Ubuntu 18.04. All the code utilized in this paper is open-source and will be made freely available in our GitHub page\footnote{\url{https://github.com/TIERS}}. Specifically for this paper we have written the following ROS packages: \textit{uwb-graph-rigidity}, \textit{uwb-collaborative-sensing} and \textit{offboard-control}. We also use \textit{dashgo-d1-ros}, \textit{ros-dwm1001-uwb-localization} and \textit{tfmini-ros} for sensor interfacing. Most of the code has been written in Python or C++. In particular, the point cloud library (PCL)~\cite{rusu20113d} is utilized to extract the position of the MAV for estimating the localization graph orientation, and extracting the point cloud around it. We use MAVROS to interface the onboard computer with the PX4 controller.

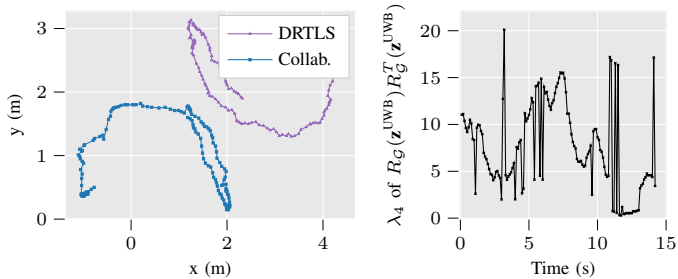
\begin{figure}
    \hspace*{-0.23in}
    \setlength{\figureheight}{0.24\textwidth}
    \setlength{\figurewidth}{0.3\textwidth} 
    \scriptsize{
\begin{tikzpicture}

\definecolor{color1}{rgb}{0.12156862745098,0.466666666666667,0.705882352941177}
\definecolor{color0}{rgb}{0.580392156862745,0.403921568627451,0.741176470588235}

\begin{axis}[
    height=\figureheight,
    width=\figurewidth,
    axis background/.style={fill=white!91!black},
    axis line style={white},
    legend cell align={left},
    legend style={draw=white!80.0!black},
    tick align=outside,
    tick pos=left,
    x grid style={white},
    xlabel={x (m)},
    xmajorgrids,
    xmin=-1.375, xmax=4.675,
    xtick style={color=black},
    y grid style={white},
    ylabel={y (m)},
    ymajorgrids,
    ymin=0.000999999999999918, ymax=3.279,
    ytick style={color=black},
]
\addplot [thin, color0, mark=triangle*, mark size=0.42, mark options={solid}]
table {%
2.32 1.9
2.29 1.92
2.24 1.97
2.2 1.99
2.15 2.01
2.09 2.06
2.03 2.05
2 2.11
1.98 2.15
1.97 2.18
1.94 2.26
1.91 2.32
1.85 2.39
1.81 2.46
1.74 2.48
1.68 2.5
1.68 2.57
1.64 2.66
1.64 2.7
1.61 2.75
1.55 2.86
1.51 2.92
1.48 2.9
1.46 2.94
1.43 2.98
1.39 2.98
1.37 3
1.36 3.01
1.34 3.03
1.3 3.05
1.29 3.07
1.3 3.08
1.21 3.1
1.25 3.13
1.21 3.05
1.2 2.99
1.19 3.01
1.25 2.9
1.22 2.85
1.24 2.86
1.29 2.8
1.31 2.82
1.33 2.81
1.33 2.77
1.3 2.77
1.3 2.75
1.3 2.76
1.29 2.72
1.27 2.7
1.27 2.72
1.28 2.7
1.29 2.69
1.27 2.61
1.3 2.58
1.36 2.45
1.41 2.38
1.43 2.32
1.47 2.26
1.5 2.2
1.54 2.13
1.56 2.1
1.59 2.05
1.62 2
1.6 1.98
1.64 1.94
1.69 1.93
1.76 1.89
1.82 1.84
1.88 1.78
1.97 1.76
2.03 1.71
2.1 1.72
2.16 1.66
2.2 1.55
2.22 1.54
2.26 1.52
2.31 1.5
2.37 1.51
2.45 1.5
2.51 1.52
2.6 1.53
2.67 1.48
2.76 1.47
2.84 1.43
2.9 1.4
2.97 1.42
3.03 1.37
3.09 1.31
3.15 1.33
3.24 1.31
3.28 1.34
3.34 1.3
3.41 1.31
3.49 1.34
3.54 1.41
3.65 1.45
3.73 1.5
3.74 1.49
3.79 1.51
3.84 1.52
3.9 1.51
3.94 1.52
3.98 1.58
4.02 1.63
4.02 1.68
4.04 1.71
4.08 1.74
4.12 1.8
4.13 1.84
4.14 1.84
4.16 1.88
4.1 1.94
4.12 2.03
4.15 2.11
4.2 2.17
4.22 2.26
4.2 2.35
4.21 2.48
4.24 2.48
4.24 2.59
4.26 2.6
4.26 2.61
4.27 2.64
4.34 2.65
4.38 2.66
4.37 2.66
4.37 2.72
4.35 2.75
4.36 2.76
4.4 2.75
4.38 2.78
4.37 2.79
4.37 2.79
4.37 2.75
4.38 2.71
4.35 2.7
4.35 2.68
4.36 2.66
4.35 2.65
4.32 2.65
4.27 2.64
4.24 2.63
4.2 2.62
4.16 2.64
4.11 2.61
};
\addlegendentry{DRTLS}
\addplot [thin, color1, mark=square*, mark size=0.42, mark options={solid}]
table {%
1.18 1.6
1.23 1.58
1.27 1.55
1.29 1.49
1.3 1.44
1.33 1.42
1.35 1.37
1.4 1.34
1.45 1.27
1.47 1.15
1.49 1.04
1.51 0.95
1.55 0.88
1.61 0.84
1.66 0.82
1.7 0.73
1.72 0.66
1.76 0.62
1.78 0.57
1.8 0.52
1.82 0.45
1.85 0.4
1.88 0.38
1.9 0.33
1.93 0.3
1.95 0.24
1.97 0.23
1.98 0.21
2 0.19
2.01 0.18
2.03 0.19
2.06 0.21
1.99 0.15
2.01 0.15
2.03 0.15
2.05 0.19
2.05 0.24
2.05 0.23
2.05 0.27
2.04 0.33
2.03 0.38
2.01 0.39
2 0.44
1.98 0.44
1.98 0.46
1.98 0.48
1.98 0.55
1.92 0.54
1.89 0.53
1.94 0.62
1.96 0.66
1.98 0.75
1.95 0.79
1.89 0.83
1.85 0.89
1.79 0.96
1.78 1.06
1.76 1.15
1.71 1.2
1.7 1.28
1.68 1.32
1.65 1.39
1.62 1.42
1.58 1.45
1.53 1.46
1.5 1.5
1.46 1.53
1.41 1.59
1.33 1.62
1.26 1.65
1.24 1.7
1.23 1.73
1.21 1.73
1.21 1.75
1.19 1.78
1.17 1.77
1.11 1.67
1.05 1.69
0.99 1.71
0.91 1.71
0.82 1.73
0.71 1.76
0.58 1.75
0.43 1.76
0.34 1.77
0.24 1.78
0.2 1.82
0.1 1.8
0.0299999999999998 1.8
-0.0700000000000003 1.8
-0.13 1.8
-0.21 1.79
-0.26 1.77
-0.32 1.76
-0.36 1.75
-0.39 1.73
-0.42 1.7
-0.43 1.62
-0.47 1.57
-0.52 1.41
-0.57 1.26
-0.59 1.32
-0.65 1.3
-0.76 1.27
-0.79 1.24
-0.97 1.17
-0.98 1.16
-1.07 1.1
-1.02 1.08
-1.02 1.08
-1.03 1.07
-1.1 1.01
-1.07 0.99
-1.06 0.95
-1.02 0.91
-0.99 0.88
-1.05 0.78
-1.02 0.71
-1.03 0.64
-1.03 0.53
-1.04 0.39
-1.04 0.39
-1.03 0.39
-1.01 0.39
-1 0.4
-0.99 0.39
-1 0.38
-1.03 0.37
-1.02 0.37
-1.01 0.36
-1 0.36
-0.97 0.37
-0.94 0.39
-0.93 0.39
-0.91 0.39
-0.9 0.4
-0.91 0.42
-0.89 0.43
-0.9 0.43
-0.92 0.42
-0.93 0.42
-0.96 0.41
-0.94 0.42
-0.830000000000001 0.47
-0.77 0.5
};
\addlegendentry{Collab.}
\end{axis}

\end{tikzpicture}}
    \setlength{\figureheight}{0.24\textwidth}
    \setlength{\figurewidth}{0.24\textwidth} 
    \scriptsize{
\begin{tikzpicture}

\begin{axis}[
    height=\figureheight,
    width=\figurewidth,
    axis background/.style={fill=white!91!black},
    axis line style={white},
    tick align=outside,
    tick pos=left,
    x grid style={white},
    xlabel={Time (s)},
    xmajorgrids,
    xmin=-0.1, xmax=15.1,
    xtick style={color=black},
    y grid style={white},
    ylabel={$\lambda_4$ of $R_{\mathcal{G}}(\textbf{z}^{\text{\tiny UWB}})R_{\mathcal{G}}^T(\textbf{z}^{\text{\tiny UWB}})$},
    ymajorgrids,
    ymin=-0.1, 
    ytick style={color=black}
]
\addplot [thin, black, mark=x, mark size=0.6, mark options={solid}]
table {%
0.1 11.0478223601051
0.2 11.1089062641868
0.3 10.3859982821927
0.4 9.61931663340136
0.5 9.19813569412637
0.6 9.69638586396975
0.7 10.4931905068403
0.8 10.0977814659899
0.9 8.44002497143339
1 8.32863196339913
1.1 2.60682619130644
1.2 9.66196288960714
1.3 9.9687020960659
1.4 9.89991523758721
1.5 9.29166948139869
1.6 8.96139104637509
1.7 6.96626961829533
1.8 6.57076403232897
1.9 6.29244430618578
2 5.77946427752031
2.1 5.23507177450167
2.2 4.69495140807464
2.3 4.61511651405434
2.4 4.04625372462173
2.5 4.35924140771734
2.6 4.92711047019344
2.7 5.03295485610579
2.8 4.54812264287655
2.9 4.29984807574317
3 1.99522943460176
3.1 12.7268925704041
3.2 20.1069326555198
3.3 4.58640970046222
3.4 4.09710532413419
3.5 4.44794959432591
3.6 4.66857834555023
3.7 4.87965824715175
3.8 5.32842029356328
3.9 5.85814928335332
4 2.00728317622851
4.1 7.58310314881736
4.2 7.41594208490442
4.3 8.09582519256823
4.4 8.33821860346039
4.5 2.67182377440886
4.6 3.15309051867157
4.7 11.2087942024938
4.8 10.3371171676863
4.9 10.6515648747078
5 11.0080062428054
5.1 11.6150259688559
5.2 12.8009926708249
5.3 12.3553786297615
5.4 4.1091371353406
5.5 14.0698027726783
5.6 14.1229171852082
5.7 14.4436214167263
5.8 4.52360609997891
5.9 14.8826268787659
6 4.11108337157918
6.1 13.9978898042081
6.2 13.3111844294957
6.3 13.4728235271728
6.4 12.6482700875689
6.5 11.9812543508928
6.6 11.5699437748363
6.7 12.285577428244
6.8 12.5883872675783
6.9 13.3498901111666
7 14.0144995355324
7.1 14.4027457948356
7.2 14.9600090172054
7.3 15.5438046231887
7.4 15.4462270770402
7.5 15.5189562481969
7.6 14.9097092419524
7.7 13.4401113564387
7.8 11.149643134103
7.9 11.149643134103
8 10.1791607554798
8.1 8.95000751798872
8.2 8.12220091023381
8.3 7.69580753000498
8.4 7.43827850538947
8.5 6.30577551213096
8.6 6.04045443293383
8.7 6.01111210374776
8.8 6.1103696904144
8.9 5.70476681185807
9 5.49759163030881
9.1 5.60164882068057
9.2 6.45185579536381
9.3 6.84494444081288
9.4 7.10881350585978
9.5 7.7286201865936
9.6 2.49402697862664
9.7 9.27187337170398
9.8 9.49714022845664
9.9 9.49714022845664
10 8.61698388162228
10.1 8.3579486202098
10.2 7.26266534390319
10.3 7.09443453157012
10.4 6.45961347642185
10.5 5.48097497590961
10.6 4.63946056121535
10.7 4.40139444621363
10.8 4.52045706481033
10.9 17.1648376693098
11 16.8288707006427
11.1 0.744138831950934
11.2 0.694431697818767
11.3 16.5273965684771
11.4 0.561621439213322
11.5 16.3453322662306
11.6 0.324408565525903
11.7 0.278050982759657
11.8 1.15740711079917
11.9 0.420761240301872
12 0.476992933965817
12.1 0.527120148587424
12.2 0.526253361686173
12.3 0.528567409154624
12.4 0.530461320045675
12.5 0.539530313562194
12.6 0.724114827896598
12.7 0.724114827896598
12.8 0.725038384638198
12.9 0.80084060013586
13 0.839055814670682
13.1 3.1818675599456
13.2 3.48173084321359
13.3 3.66429652032554
13.4 4.1267809472976
13.5 4.35763289552468
13.6 4.75939319022881
13.7 4.56887210459617
13.8 4.57834357237237
13.9 4.57515312572851
14 4.38051873121881
14.1 17.1383992557708
14.2 3.4409148471087
};
\end{axis}

\end{tikzpicture}}
    \caption{(Left) Localization based on Decawave's fixed-anchor DRTLS and the collaborative localization framework (\textit{collab}), using the UGV's position as the origin of coordinates, and therefore giving relative localization only (rototranslation from DRTLS except for measurement errors). (Right) The rigidity eigenvalue is monitored during flight to ensure that the graph is always rigid.}
    \label{fig:rigidity}
\end{figure}

\begin{figure*}
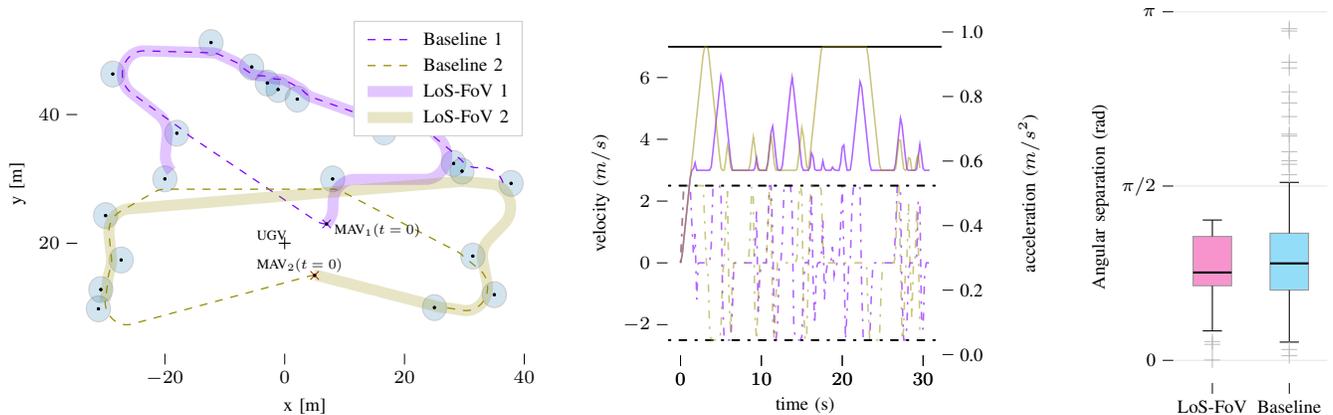

    \centering
    \begin{subfigure}[t]{0.42\textwidth}
        \setlength{\figureheight}{0.8\textwidth}
        \setlength{\figurewidth}{\textwidth} 
        \scriptsize{\input{tex_final/paths}}
        \caption{Solution to the DMTSPN (MAV paths). The neighborhoods are shown in blue, representing the locations to be visited.}
        \label{fig:paths_comparison}
    \end{subfigure}
    \begin{subfigure}[t]{0.36\textwidth}
        \setlength{\figureheight}{0.9\textwidth}
        \setlength{\figurewidth}{0.8\textwidth} 
        \scriptsize{\input{tex_final/speeds}}
        \caption{Velocity and acceleration profiles for each of the MAVs (solution with LoS-FoV constraints only).}
        \label{fig:profiles}
    \end{subfigure}
    \begin{subfigure}[t]{0.2\textwidth}
        \setlength{\figureheight}{1.8\textwidth}
        \setlength{\figurewidth}{\textwidth} 
        \scriptsize{
\begin{tikzpicture}

\definecolor{color1}{rgb}{0.73,0.17,0.03}

\definecolor{color0}{rgb}{0.5,0,1}

\begin{axis}[
    height=\figureheight,
    width=\figurewidth,
    axis line style={white},
    legend style={fill opacity=0.8, draw opacity=1, text opacity=1, draw=white!80!black},
    tick align=outside,
    tick pos=left,
    x grid style={white!69.0196078431373!black},
    xmin=0.5, xmax=2.5,
    xtick style={color=black},
    xtick={1,2},
    xticklabels={LoS-FoV, Baseline},
    y grid style={white!90!black},
    ylabel={Angular separation (rad)},
    ymajorgrids,
    ymin=-0.2, ymax=3.15,
    ytick style={color=black},
    ytick={0,1.57,3.14},
    yticklabels={0,$\pi/2$,$\pi$},
    scaled y ticks = false
]
\addplot [black, forget plot]
table {%
1 0.671888665909249
1 0.26519936569698
};
\addplot [black, forget plot]
table {%
1 1.11607447169562
1 1.2639975677668
};
\addplot [black, forget plot]
table {%
0.875 0.26519936569698
1.125 0.26519936569698
};
\addplot [black, forget plot]
table {%
0.875 1.2639975677668
1.125 1.2639975677668
};
\addplot [lightgray, mark=+, mark size=3, mark options={solid}, only marks, forget plot]
table {%
1 0.1688171869247
1 0.00275188045108998
1 0.14367841299196
};
\addplot [black, forget plot]
table {%
2 0.632770822752011
2 0.16415703345617
};
\addplot [black, forget plot]
table {%
2 1.14526621803757
2 1.60324907120309
};
\addplot [black, forget plot]
table {%
1.875 0.16415703345617
2.125 0.16415703345617
};
\addplot [black, forget plot]
table {%
1.875 1.60324907120309
2.125 1.60324907120309
};
\addplot [lightgray, mark=+, mark size=3, mark options={solid}, only marks, forget plot]
table {%
2 0.09727988645249
2 0.0417812914883098
2 1.79833233491784
2 2.03697058211828
2 2.31836279451422
2 2.63389514273772
2 2.96658177569715
2 2.98863399239257
2 2.68512883056327
2 2.41810356514095
2 2.1958462749282
2 2.01983458246078
2 1.87936338617503
2 1.76570675644667
2 1.67241794664869
};
\path [draw=black, fill=magenta, opacity=0.42]
(axis cs:0.75,0.671888665909249)
--(axis cs:1.25,0.671888665909249)
--(axis cs:1.25,1.11607447169562)
--(axis cs:0.75,1.11607447169562)
--(axis cs:0.75,0.671888665909249)
--cycle;
\path [draw=black, fill=cyan, opacity=0.42]
(axis cs:1.75,0.632770822752011)
--(axis cs:2.25,0.632770822752011)
--(axis cs:2.25,1.14526621803757)
--(axis cs:1.75,1.14526621803757)
--(axis cs:1.75,0.632770822752011)
--cycle;
\addplot [thick, black, forget plot]
table {%
0.75 0.791403111994706
1.25 0.791403111994706
};
\addplot [thick, black, forget plot]
table {%
1.75 0.873271293432525
2.25 0.873271293432525
};
\end{axis}

\end{tikzpicture}}
        \caption{Distribution of angular distance between MAVs.}
        \label{fig:boxplot}
    \end{subfigure}
    \caption{Simulations for two MAVs and one UGV with limited FoV. Subfigure (a) shows both a baseline DMTSPN solver, and our solver with LoS constraints, with (b) showing the velicity and acceleration profiles of our solver. This particular example was chosen to illustrate how the MAVs follow an anticlockwise direction from the UGV's reference.
    In Subfigure (c), we show that only with our DMTSPN solver we can ensure that MAVs are always within LoS with limited FoV at the UGV (angular separation always below a predefined FoV of $\pi/2$).}
    \label{fig:dmtspn}
    \vspace{-1em}
\end{figure*}

\begin{figure}
    \centering
    \begin{subfigure}[t]{0.23\textwidth}
        \centering
        \includegraphics[width=\textwidth,height=0.8\textwidth]{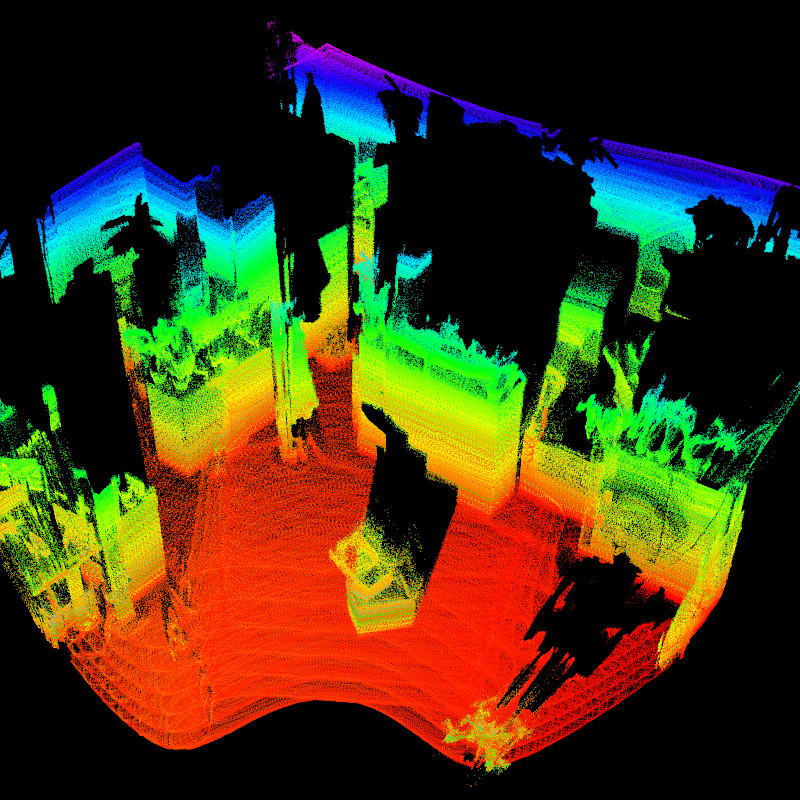}
        \caption{Global scene with boxes in the center and MAV in the bottom right.}
        \label{fig:global_scan}
    \end{subfigure}
    \hfill
    \begin{subfigure}[t]{0.23\textwidth}
        \centering
        \includegraphics[width=\textwidth,height=0.8\textwidth]{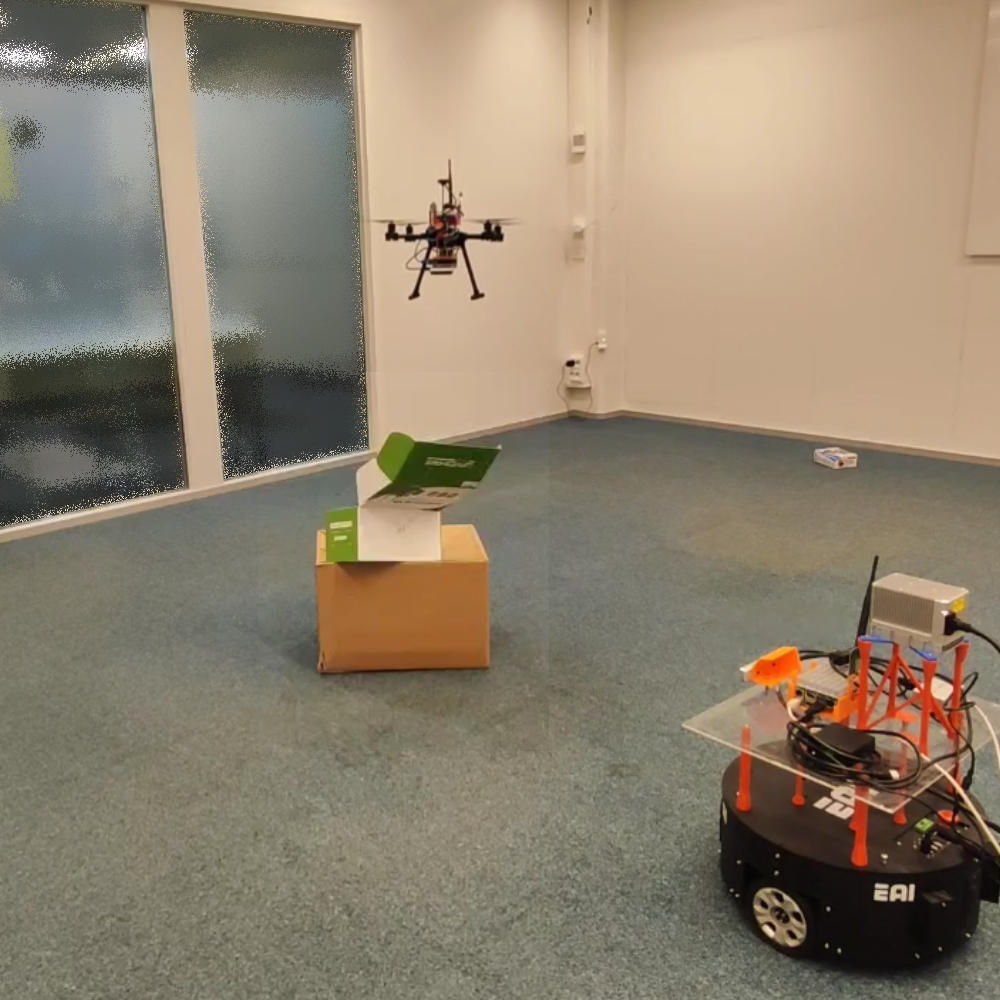}
        \caption{Picture of the test environment while the MAV is hovering.}
    \end{subfigure}
    
    \vspace{0.42em}
    
    \begin{subfigure}[t]{\linewidth}
        \centering
        \includegraphics[width=\textwidth,height=0.4\textwidth]{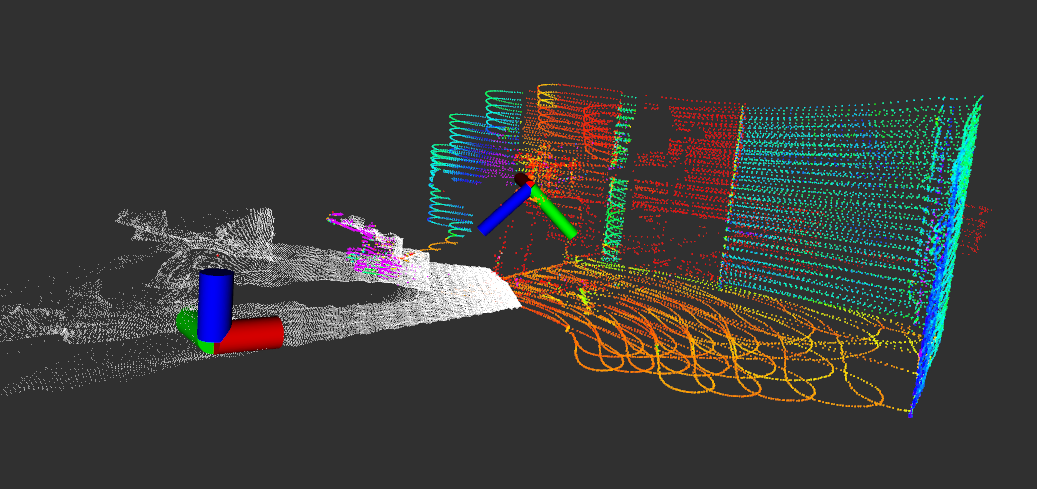}
        \caption{Aligned point clouds using UWB+VIO for relative localization. The white point cloud is recorded with the RealSense D435 depth camera on the MAV, while the colored point cloud comes from the Livox lidar on the UGV (both poses shown, D345 is inclined $36\degree$).} 
    \end{subfigure}
    \caption{Validation of the collaborative sensing algorithm. Figures (a) and (b) show the scene from opposite corners. The UGV is not visible in (a). 
    Figure (c) shows the limited FoV of the Livox lidar.}
    \label{fig:reconstruction}
\end{figure}

\section{Experimental Results}\label{sec:experimental_results}

This section reports the results from the simulations and experiments carried out to validate the proposed methods. First, we perform test flights to assess the viability of the collaborative localization framework, using the UGV and MAV and additional UWB transceivers on the ground that are also considered as graph vertexes to ensure rigidity. Second, we introduce simulations to show that an UGV is able to track and maintain in its FoV two MAVs while they are exploring the occluded locations around the UGV (not possible with a basic DMTSPN solver). Then, we show that we achieve good alignment of raw data collected from the UGV and MAV 
for 3D reconstruction. 
Finally, we discuss the system's scalability.

\textbf{Collaborative Localization Framework.} Fig.~\ref{fig:rigidity} shows the positions of the MAV based on Decawave's anchor-based localization system (DRTLS) and our collaborative localization framework over a test flight. The UGV is taken as the origin of coordinates in the latter scenario. We also show the evolution of the rigidity eigenvalue. Because our framework provides relative localization only, the two paths in the figure are congruent with respect to a rototranslation, except for measurement errors. However, by detecting the MAV from the UGV after take-off, we are able to fix the orientation of the localization graph in the common orientation frame (not necessarily the same than the DRTLS system). The DRTLS localization is based on 8 fixed anchors, and therefore the accuracy can be considered higher. Monitoring the rigidity eigenvalue will play a more important role in applications where the robots operate in a larger environment, and this is, to the best of our knowledge, the first time it is used as a \textit{health} indicator to the system for higher-level planning.

\textbf{Multi-MAV Path Planning.} The multi-robot path planning with UGV-to-MAV LoS constraints in limited FoV scenarios is tested through simulations with random distribution of neighborhoods. The simulations are done with a fixed UGV that can only rotate and two MAVs that have to visit all locations in the map as fast as possible while staying in LoS within the FoV of the UGV. We pick one representative example, and compare in Fig.~\ref{fig:reconstruction} the difference between the paths calculated by a baseline DMTSPN without constraints, and ours. Fig.~\ref{fig:boxplot} shows the angular distance between the two MAVs from the UGV's referece when considering polar coordinates. For the solver with constraints, the FoV is limited to $\pi/2$~rad. We can see that there are multiple times where the UGV would be unable to maintain both MAVs within its FoV (points where the angular distance is larger than $\pi/2$~rad) with the baseline DMTSPN implementation. By introducing the limited FoV constraints, we are therefore able to obtain results that can be better ported to real-world applications.



\textbf{Collaborative Scene Reconstruction.} We perform experiments in an indoors facility and show the performance of the collaborative localization and sensing algorithms for matching raw point clouds (see Fig.~\ref{fig:reconstruction}). First, the UGV scans the scene. To create a map, we utilize an implementation of lidar odometry and mapping (LOAM) optimized for limited-FoV lidars: Livox LOAM~\cite{lin2020loam}. We then detect the blind spots to the UGV based on a standard elevation occupancy map. In the experiments, owing to the limited space available, we only use one UGV and one MAV, together with two more UWB transceivers placed on the ground to ensure global graph rigidity. The UGV is set to rotate and move within a limited space to always maintain the MAV within its FoV and in LoS. Fig.~\ref{fig:reconstruction} shows the aligned point clouds of the MAV and UGV, together with their poses in a given instant. The colored point cloud is obtained from the Livox lidar, with the non-repetitive scan pattern being recognizable by the \textit{waves} in the ground. Two stacked boxes in the middle of the test area are scanned from complementary points of view simultaneously, showing a good point cloud alignment directly with the raw data. This opens the door to further optimization, and providing feedback to the localization framework based on the alignment of local point clouds. The area behind the boxes occluded to the UGV's lidar corresponds to a neighborhood in the DMTSPN formulation. Intuitively, the neighborhoods can be mapped to the UGV lidar'\textit{shadows} as those seen in Fig.~\ref{fig:global_scan}.

\textbf{Scalability.} In terms of the system's scalability, the localization framework is based on UWB ranging and the VIO egomotion estimation is done on a separate processor on the T265 camera. The UWB ranging is only limited by the number of robots in terms of the available bandwidth, and therefore the localization frequency must be decreased as the number of robots increases (the current frequency of 10\,Hz can accommodate approximately 20 robots within line-of-sight of each other). 
As to the number of robots involved in scene reconstruction, because the MAVs are assigned to one UGV and the DMTSPN solved from each UGV's reference, the computational load of the path planning algorithm can be maintained even when the number of robots grows. The assignment of MAVs to UGVs has linear complexity.

In summary, we provide an initial implementation of all parts of the proposed system, which can then be leveraged for collaborative scene reconstruction in GNSS-denied environments. In particular, we show that the localization framework has potential to take multi-robot systems out of the lab with good results in local map merging for scene reconstruction, lifting the need for more accurate but significantly less flexible motion capture systems or UWB localization systems based on fixed anchors. Finally, the fact that UGVs are continuously tracking MAVs and keeping them within their FoV in LoS means that the more accurate lidar data they capture can be used for local path planning on the MAVs. This opens the door to operating the MAVs in more complex and dynamic environments even with limited onboard sensing. 

\section{Conclusion}\label{sec:conclusions}

We have addressed some of the challenges in multi-robot dense scene reconstruction, with a focus on (i) collaborative localization, and (ii) path planning with LoS and FoV constraints. We have first presented a framework for collaborative localization with UWB-based ranging and VIO fusion in heterogeneous UGV+MAV systems, exploiting sensors onboard the UGVs to establish a common reference frame for all agents. Then, we have utilized this framework for collaborative scene reconstruction.
With simulations, we show that the proposed algorithm effectively ensures MAVs are always within the UGVs FoV, and with real experiments we show that the localization framework is accurate enough to provide good alignment of point clouds even at the raw data level.

In future works, we will look into improving the DMTSPN solution by allowing dynamic tracking of a single MAV from different UGVs during the scene reconstruction mission. We will also investigate the possibilities of integrating the navigation of the UGVs within the DMTSPN solver.


\section*{Acknowledgment}

This research work is supported by the Academy of Finland's AutoSOS and RoboMesh projects (Grant No. 328755) and the Swiss National Science Foundation with grant number 200020 188457.

\bibliographystyle{IEEEtran}
\bibliography{bibliography}

\end{document}